\documentclass[letterpaper, 10 pt, conference]{ieeeconf}  % Comment this line out if you need a4paper

\IEEEoverridecommandlockouts                              % This command is only needed if 
\usepackage{amsmath} 
\usepackage{amssymb}
\usepackage[mathscr]{euscript}
\usepackage{graphicx}
\usepackage[table]{xcolor}
\usepackage{algorithm}
\usepackage[noend]{algpseudocode}
\usepackage{tabularx}
\usepackage{multirow}
\usepackage{hhline}
\usepackage{hyperref}
\usepackage[T1]{fontenc}
\usepackage{helvet}
\usepackage{lipsum}
\usepackage{stfloats}
\usepackage{subcaption}
\usepackage{lipsum}  
\usepackage{arydshln}
\usepackage{booktabs}
\usepackage{textcomp, gensymb}
\let\labelindent\relax % needed for enumitem as labelindent is already defined
\usepackage{enumitem}

\usepackage[capitalise,nameinlink]{cleveref}
  \crefname{section}{Sect.}{Sect.}
  \Crefname{section}{Section}{Sections}
  \crefname{figure}{Fig.}{Fig.}
  \Crefname{figure}{Figure}{Figures}
  \crefname{table}{Tabl.}{Tabl.}
  \Crefname{table}{Table}{Tables}

\newcommand{\methodName}{Diffusion Policies For Compliant Manipulation}
\newcommand{\methodAbbr}{DIPCOM}

\title{\LARGE \bf
Learning Diffusion Policies from Demonstrations For Compliant Contact-rich Manipulation
}

\author{
Malek Aburub$^{1, 2^*}$, 
Cristian C. Beltran-Hernandez$^{1^*}$, 
Tatsuya Kamijo$^{1}$, 
Masashi Hamaya$^{1}$% <-this % stops a space
\thanks{$^{*}$ Equal contribution.}
\thanks{$^{1}$OMRON SINIC X Corporation, Tokyo, Japan}
\thanks{$^{2}$Department of Engineering Science, Osaka University, Japan}
\thanks{Corresponding authors:}
\thanks{\tt\small cristian.beltran [at] sinicx.com}
\thanks{This work was partly supported by the JST-Mirai Program (Grant Number JPMJMI21G2).}
}

\begin{document}
\maketitle

\begin{abstract}

Robots hold great promise for performing repetitive or hazardous tasks, but achieving human-like dexterity, especially in contact-rich and dynamic environments, remains challenging. Rigid robots, which rely on position or velocity control, often struggle with maintaining stable contact and applying consistent force in force-intensive tasks. Learning from Demonstration has emerged as a solution, but tasks requiring intricate maneuvers, such as powder grinding, present unique difficulties. This paper introduces Diffusion Policies For Compliant Manipulation (DIPCOM), a novel diffusion-based framework designed for compliant control tasks. By leveraging generative diffusion models, we develop a policy that predicts Cartesian end-effector poses and adjusts arm stiffness to maintain the necessary force. Our approach enhances force control through multimodal distribution modeling, improves the integration of diffusion policies in compliance control, and extends our previous work by demonstrating its effectiveness in real-world tasks. We present a detailed comparison between our framework and existing methods, highlighting the advantages and best practices for deploying diffusion-based compliance control.

\end{abstract}

\section{Introduction}
% Why dexterous contact-rich manipulation tasks are challenging

Robots hold significant potential to enhance daily life by performing repetitive or hazardous tasks. To achieve this, robots must be able to manipulate objects with a level of dexterity comparable to that of humans. However, enabling robots to reach human-like dexterity remains a challenging problem, especially for tasks that require precise, contact-rich manipulation in dynamic environments, such as laboratories and workshops. Rigid robots, which typically operate through position or velocity commands, face difficulties maintaining stable surface contact and applying consistent force—both essential for force-intensive tasks.

\begin{figure}[t]
    \centering
    \includegraphics[width=\linewidth]{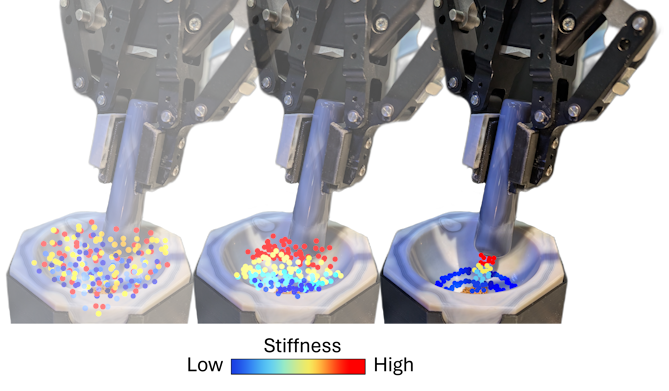}
     \caption{\textbf{DIPCOM Denoising Process:} illustration of how DIPCOM predicts end-effector positions and adjusts arm stiffness during multiple timesteps, enabling consistent force during contact and smooth transitions between movements.} 
    \label{grinding_overview}
\end{figure}

Learning from Demonstration (LfD) techniques~\cite{ravichandar2020survey} have emerged as a promising approach, enabling robots to acquire complex tasks by observing human experts and replicating their precise actions. However, tasks that require intricate maneuvers and high force, such as powder grinding in Fig. \ref{grinding_overview}, pose unique challenges. In such scenarios, rigid robots often fail to sustain controlled contact and force over extended periods, limiting their effectiveness in demanding applications. In such cases, rigid robots necessitate mechanical compliance mechanisms \cite{nakajima2022robotic} to make the physical interaction safer at the expense of making it harder to precisely control the position of the tool, i.e., the pestle.

% What are we doing to tackle the challenges
To address these challenges, we adopt compliance control schemes that enable rigid robots to handle tasks involving direct contact by modulating forces through external sensors \cite{calanca2015review}. 
In our prior work~\cite{kamijo2024learning}, a method called Comp-ACT was proposed that combines force information and compliance control with a policy based on Variational AutoEncoders and the Action Chunking with Transformers strategy (VAE-ACT). The Comp-ACT policies perform well in various tasks but struggle to solve long-horizon tasks with repetitive behaviors, such as grinding powder. Therefore, we extend compliant LfD policies by incorporating diffusion models, enabling robots to adaptively regulate force during task execution while maintaining precision and stability, especially during long-horizon tasks.
Our approach builds on the robot demonstration system for contact-rich manipulation introduced in \cite{kamijo2024learning}, improving its performance in handling force-intensive tasks.

% Overview of our proposed method/framework
In this work, we propose \methodName~(\methodAbbr), a novel diffusion-based framework that leverages generative models to handle compliant control tasks. 
Our choice of diffusion-based policies is motivated by their potential to capture multi-modal action distributions and produce a more diverse behavior than VAE-ACT-based policies~\cite{jia2024towards}.
Diffusion models progressively add noise to data in a forward stochastic process. By training a neural network known as a denoising model, they approximate the original data and iteratively generate new actions. We implement diffusion models into compliance control to develop a policy capable of predicting both the Cartesian end-effector (EE) pose and simultaneously adjusting the stiffness of the robot arm to apply the necessary force for the task.

% Summary of concrete contributions
% 1. Force-Aware Diffusion policy
% 2. Comparison with Comp-ACT
To summarize our contributions: 
\begin{itemize}
    \item A novel diffusion-based framework for rigid robots to learn complex contact-rich manipulations from demonstrations via compliance control. We show that the expressiveness of diffusion models, along with their ability to capture multimodal distributions, enhances force control.
    % \item we provide technical insights on how to effectively integrate these policies into a compliance control setting, overcoming the challenges posed by the nature of diffusion models.
    \item A comprehensive comparison of \methodAbbr~and Comp-ACT \cite{kamijo2024learning} on challenging real-world tasks, highlighting the specific tasks where diffusion-based policies excel and outlining best practices for their application.
\end{itemize}

\section{Related Works}
\subsection{Learning from Demonstrations for Contact-rich Manipulation}

Learning from demonstration has emerged as a promising approach for teaching robots complex contact-rich manipulation skills \cite{calinon2009robot}. Researchers have explored using force/torque (F/T) sensing and haptic feedback to capture expert demonstrations to learn tasks such as grasping \cite{schmidts2011imitation}, ironing \cite{kormushev2011imitation}, pouring \cite{rozo2013robot}, and peg-in-hole insertion \cite{wang2021robotic}. Recent work has focused on sample-efficient methods that can learn from a few demonstrations through transformer-based models and innovative teleoperation interfaces such as ALOHA~\cite{zhao2023learning} and Universal Manipulation Interface (UMI)~\cite{chi2024universal}. 

While most prior approaches focus on position control and mechanical compliance to allow robots a degree of safety while performing contact-rich tasks, this work focuses on active compliance control to perform contact-rich manipulation tasks.

In our prior work \cite{kamijo2024learning}, we proposed an intuitive teleoperation interface for collecting demonstrations and the Compliance Control via Action Chunking with Transformers (Comp-ACT) method to learn compliance control policies from a few demonstrations. 
The current study utilizes the same teleoperation interface to collect demonstrations for training a \methodAbbr~policy.

Similar to the work of Drolet et al. \cite{drolet2024} and Zhao et al. \cite{zhaoaloha}, where several imitation learning methods, including Action Chunking with Transformers (ACT)~\cite{zhao2023learning} and Diffusion policies~\cite{chi2023diffusion},  were compared on bimanual manipulation tasks, this study compares the performance of Comp-ACT~\cite{kamijo2024learning} with the proposed \methodAbbr~with a focus on rigid robots tackling challenging real-world contact-rich tasks. 

\subsection{Diffusion Policies}

Diffusion models, introduced by Ho et al. \cite{ho2020denoising}, are probabilistic generative models that transform random noise into meaningful samples from a target distribution. For a more comprehensive survey on diffusion models, see \cite{yang2023diffusion}.

In robotics, diffusion models have effectively captured multi-modal actions and have been applied to various domains, including motion planning \cite{carvalho2023motion}, navigation \cite{ryu2024diffusion}, human-robot interaction \cite{ng2023diffusion}, and grasping tasks \cite{urain2023se}. In manipulation, Chi et al. \cite{chi2023diffusion} demonstrated strong results in visuomotor policy learning from demonstrations using diffusion models.

Further advances include Liu et al. \cite{liu2022structdiffusion}, who leveraged language-annotated play data to enable skill acquisition, while Reuss et al. \cite{reuss2023goal} conditioned diffusion models on goal states to enhance skill learning from play data for task completion.

% Even with extensive training, robots struggle to maintain the necessary force and torque for precise control in these scenarios.

% Although previous work has primarily focused on improving learning efficiency or utilizing large, uncurated datasets, the reliance on position or velocity controllers often hinders performance in contact-rich tasks.
% To address this, we introduce a diffusion policy conditioned on force, integrated with a compliance controller. 
% this study we show that our framework can tackle challenging contact-rich tasks with fewer demonstrations compared to the reported by prior studies that do not use compliance control \cite{zhaoaloha}
% As demonstrated in our experiments, this approach significantly improves performance without the need for a large number of demonstrations.

Previous work has primarily focused on improving learning efficiency or using large, uncurated datasets, but the reliance on position or velocity controllers often limits their effectiveness in contact-rich tasks. To address this, we introduce a diffusion policy conditioned on force and integrated with a compliance controller. This study demonstrates that our framework can handle challenging contact-rich tasks with fewer demonstrations than reported in prior studies lacking compliance control \cite{zhaoaloha}. Our experiments show that this approach significantly improves task performance without extensive demonstrations.
\section{Methodology}

\begin{figure*}[ht]
    \centering
    \includegraphics[width=0.9\textwidth]{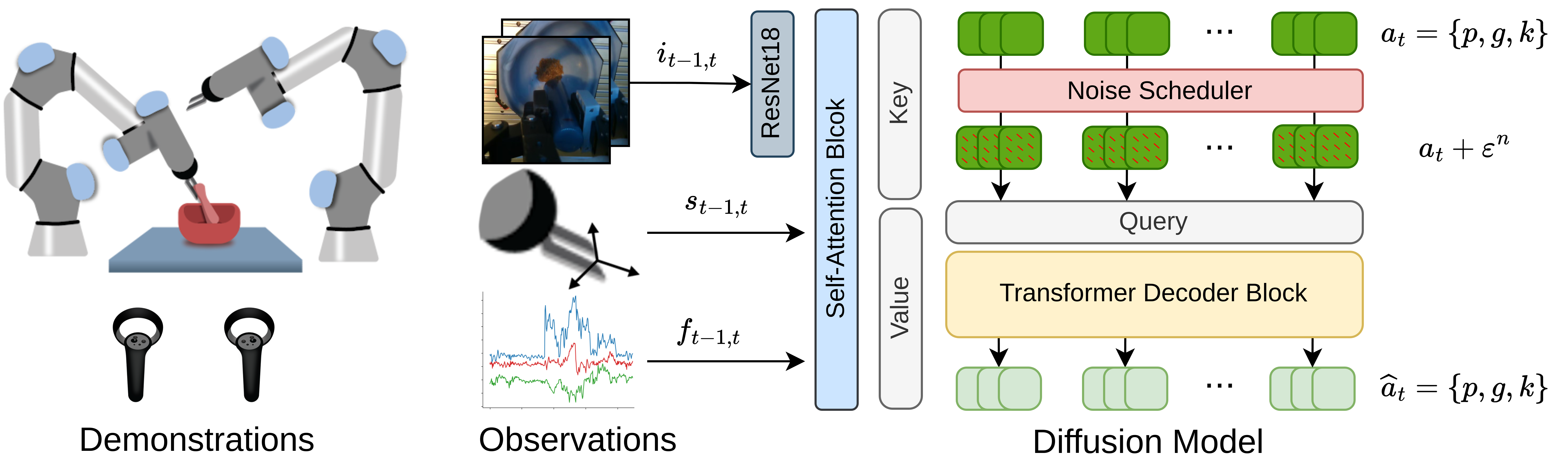} %added time enumerator to epsilon
    \caption{\textbf{Policy Framework:} Left: Dataset collection framework. Middle: Observations $O$ include images $i_{t-1,t}$, robot Cartesian pose $s_{t-1,t}$, and measured force/torque $f_{t-1,t}$, all encoded using a self-attention transformer. Right: During training, actions $a_{t}$—comprising the end-effector pose $p$, gripper pose $g$, and stiffness $K$—are processed through a noise scheduler that adds Gaussian noise $\epsilon$ over time steps $n$. These noisy actions are then input into the transformer decoder block. During inference, Gaussian noise replaces the training noise, and the transformer decoder block predicts the actions $\hat{a}_{t}$}
    \label{fig:overview}
\end{figure*}

We introduce \methodName~(\methodAbbr), a novel method for learning variable compliance control from demonstrations using diffusion models. Our approach predicts target EE poses and robot stiffness parameters conditioned on current observations, including the contact force data. A compliance controller uses these predictions to compute the final joint position commands that allow robots to move compliantly at the predicted stiffness. \cref{fig:overview} illustrates the architecture of \methodAbbr.

\subsection{Problem Formulation}

Learning from demonstration (LfD) aims to enable robots to acquire new skills by autonomously observing and imitating human-provided demonstrations. In our approach, the policy learns to predict a sequence of absolute Cartesian EE pose and stiffness parameters given current observation \(\mathcal{O}\) that include RGB images \(\mathcal{I} \in \mathbb{R}^{H \times W \times 3}\), the latest F/T sensor reading \(\mathcal{F} \in \mathbb{R}^{6}\), and proprioception data \(\mathcal{S} \in \mathbb{R}^{9}\). 
% The predicted action \(\mathcal{A}\) is represented by:
% \[
% \mathcal{A} = \{\mathbf{p}, \mathbf{g}, \mathbf{k}\}
% \]
The predicted action $\mathcal{A} = \{\mathbf{p}, \mathbf{g}, \mathbf{k}\}$ comprises three components: the absolute EE pose, the gripper action, and the stiffness parameter. The absolute EE pose, denoted as $\mathbf{p} = \{\mathbf{r}, \mathbf{o}\} \in \mathbb{R}^{9}$, consists of a position vector $\mathbf{r} \in \mathbb{R}^{3}$ and an orientation vector \(\mathbf{o} \in \mathbb{R}^{6}\). For more details on this 6D orientation representation, refer to \cref{subsec:orientation-representation}. The gripper action \(\mathbf{g} \in \mathbb{R}^{1}\) represents the desired non-binary opening width of the gripper. Finally, the stiffness parameter \(\mathbf{k} \in \mathbb{R}^{6}\) corresponds to the diagonal elements of the stiffness matrix.
Each robot action \(\mathcal{A}\) is then fed into a compliance controller.

% where \(\mathbf{p} = (\mathbf{r}, \mathbf{o})\) includes a position \(\mathbf{r} \in \mathbb{R}^{3}\) and an orientation \(\mathbf{o} \in \mathbb{R}^{6}\) (see \cref{subsec:orientation-representation} for more details), , and a stiffness vector \(\mathbf{k} \in \mathbb{R}^{6}\).

% Can we add anything to this? or we just say we introduced it in previous work 
\subsection{Data Collection}
The teleoperation system presented in \cite{kamijo2024learning} was used for data collection. The system uses Virtual Reality (VR) controllers to provide the reference EE pose and stiffness that are input to the compliance controller. During the demonstration, the operator can switch between two pre-selected stiffness modes using the grip button on the side of the controller. The action $\mathcal{A}$ from the VR controllers and the observation $\mathcal{O}$ from the robots and cameras are collected and stored.
% end-effector position and orientation, the F/T sensor readings, and the stiffness parameter of the compliance controller. 
% Additionally, RGB images were collected from multiple points of view. 

\subsection{Orientation Representation for Cartesian Action Space}\label{subsec:orientation-representation}
As in our previous work \cite{kamijo2024learning}, the robot's state observations and actions were defined as absolute Cartesian poses. In \cite{kamijo2024learning}, orientations were represented using axis angles (rotation vectors). However, this orientation representation has two problematic characteristics. The first issue is that many rotation vectors can represent the same orientation. 
The challenge with this one-to-many representation is that a model would need to be trained on all possible representations so that it can yield consistent outputs at inference time, given any of the possible representations.
The second issue is that the axis angle representation is discontinuous. Zhou et al. \cite{zhou2019continuity} discussed the discontinuity problem and concluded that neural networks can better fit continuous representations. They proposed 5D and 6D continuous orientation representations. To address both issues, in this work, we adopted the six-dimensional orientation representation proposed by \cite{zhou2019continuity}, which provides a continuous and unique representation for each orientation. The 6D orientation representation consists of the rotation matrix's first two columns. Please refer to \cite{zhou2019continuity} for a more comprehensive discussion.

\subsection{\methodName}

 Our dataset is inherently multi-modal, containing a variety of observations and the corresponding actions that must be predicted. To address this, we aim to learn a policy distribution \(\pi(\mathcal{A} | I, F, S)\) from a task-specific demonstration dataset. We introduce \methodName, a classifier-free conditional diffusion model designed to generate actions \(\mathcal{A}\) based on observations \(\mathcal{O}\).

The diffusion process involves two stages: a forward process that incrementally adds noise to the data and a learnable inverse diffusion process that recovers the original data from noise conditioned on the input observations. The forward process progressively corrupts the data, while the inverse process, learned by the model, removes the noise step by step.

We implement the Denoising Diffusion Implicit Model (DDIM) formulation from Song et al. \cite{song2021denoising} for noise scheduling and denoising. This approach makes the denoising process deterministic, facilitating more efficient inference and allowing for flexible adjustment of inference steps to optimize performance.

The architecture of DIPCOM is shown in Fig.\ref{fig:overview}. It follows the design of \cite{chi2023diffusion}, using an encoder-decoder transformer \cite{vaswani2017attention} with a ResNet18 vision backbone (without pre-trained weights). Images are processed through the vision backbone, concatenated with force and robot state data, and input into the transformer's encoder. Cross-attention is applied to the noisy actions in the decoder.

The model is trained to predict actions \(\hat{a}_0\), using the mean squared error loss function:
\[
L_{\text{sample}} = \| {a}_t^{0} - \hat{a}_t^{0} \|^2 
\]

During inference, the model iteratively predicts the original sample using the formula:
\[
a_{t}^{n-1} = \sqrt{\beta_{n-1}} \hat{a}_t^{0} + \sqrt{1 - \beta_{n-1}} \cdot \frac{a_t^{n} - \sqrt{\beta} \hat{a}_t^{n}}{\sqrt{1 - \beta}}
\]
where \(a_t^{n}\) represents the noisy data at time step \(n\), \(\beta\) is the cumulative product of noise scales up to time \(n\), and \(\hat{a}_t^{0}\) is the estimated original data.

\subsection{Action Sequence Generation} 

Applying diffusion models to contact-rich manipulation presents unique challenges due to the robot’s prolonged interaction with the object’s surface. In such tasks, variable forces are exerted, and any inconsistencies in action prediction can negatively impact performance. Previous works \cite{chi2023diffusion, chi2024universal} typically predict a fixed number of actions, using part of the horizon while discarding the rest. This can lead to instability, such as jerky or abrupt movements, especially during transitions between inference steps. While this might be acceptable for simpler tasks like pick-and-place or rearranging, it becomes problematic for contact-rich manipulation.

To address this, we operate at a higher action prediction frequency than previous works, which typically cap predictions at around 16 actions on the horizon and run at a maximum of 20 Hz. In contrast, we predict an average of 48 actions per horizon, with some tasks requiring even longer sequences, allowing the robot to handle fine-grained control demands in contact-rich scenarios.

Another challenge in using diffusion over longer horizons is the accumulation of error, especially when only a portion of the predicted actions is applied, leading to potential instability. To mitigate this, we utilize the Temporal Ensemble introduced in \cite{zhao2023learning} for ACT policy and extend to diffusion, applying the rest of the horizon to the next action steps from the next inference, smoothing transitions between action sequences and minimizing erratic movements, ensuring more stable performance in contact-rich tasks.

\section{Experiments}
\begin{figure*}[t!]
    \centering
    \captionsetup{justification=centering}
    \includegraphics[width=0.95\linewidth]{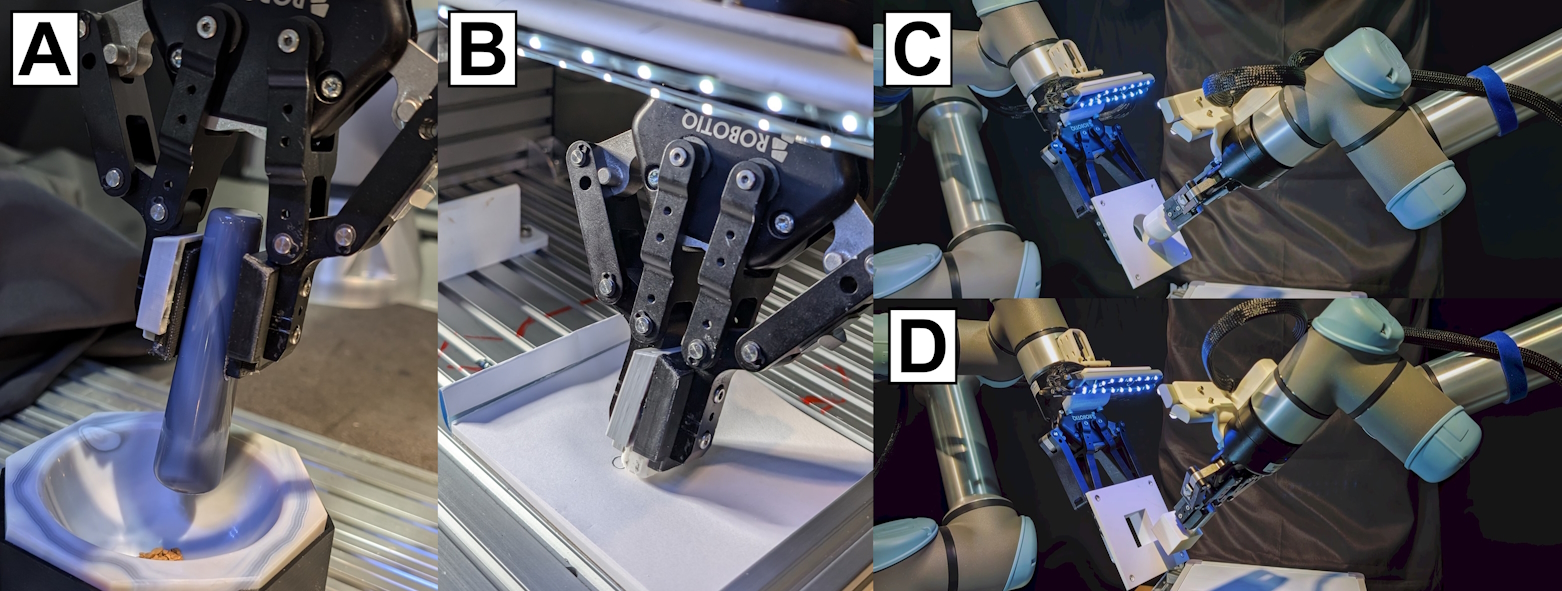}
    \caption{Contact-rich manipulation tasks used for evaluation. A - Powder grinding. B - Pencil eraser. \\ C - Bimanual round peg insertion. D - Bimanual cuboid peg insertion.
    }
    \label{fig:tasks_description}
\end{figure*}

The proposed method \methodAbbr~was evaluated on challenging contact-rich tasks that require applying force carefully and consistently to solve them. The tasks are described in \cref{sec:task_description} and illustrated in \cref{fig:tasks_description}. Additionally, we compare the performance of the proposed method against a baseline described below.

\subsection{Baseline Method}
The prior work, Comp-ACT~\cite{kamijo2024learning}, was used as the baseline method. The baseline policy is trained as the decoder of a conditional variational autoencoder (CVAE). It consists of the transformer encoder, which synthesizes all the observational data, and the transformer decoder, which generates a sequence of action, namely Cartesian EE pose and stiffness parameters. Similarly to our approach, the baseline uses F/T observations, Cartesian EE poses, and RGB images to predict actions that are passed to the compliance controller.

\subsection{Experimental Setup}
The robotic system used for experimentation consisted of two UR5e robot arms (Universal Robots A/S, 2024) with built-in F/T sensors on their wrists. Two cameras, RealSense SR305  (Intel Corporation, 2024), are attached to the wrist of each arm. Another static third-person view camera is placed in front of the robots to capture a wider field of view. 

The demonstration data from each task was collected from three co-authors to add variety to the datasets.

\subsection{Tasks description and results} \label{sec:task_description}
This section describes the tasks used for experimentation alongside the results obtained after training one policy per task for each model. The general conditions considered for each task are reported in \Cref{table:task_conditions}, such as the number of demonstrations and stiffness modes.

% Please add the following required packages to your document preamble:
% \usepackage{multirow}
\begin{table}[t]
\centering
\caption{Task conditions}
\label{table:task_conditions}
\begin{tabular}{|c|ccccccc|}
\hline
\multirow{4}{*}{\textbf{\begin{tabular}[c]{@{}c@{}}Task \end{tabular}}} &
  \multicolumn{7}{c|}{\textbf{Task Conditions}} \\ \cline{2-8} 
 &
  \multicolumn{1}{c|}{\multirow{3}{*}{\textbf{\begin{tabular}[c]{@{}c@{}}\#\\ of \\ demos.\end{tabular}}}} &
  \multicolumn{1}{c|}{\multirow{3}{*}{\textbf{\begin{tabular}[c]{@{}c@{}}\#\\ of \\ views\end{tabular}}}} &
  \multicolumn{1}{c|}{\multirow{3}{*}{\textbf{\begin{tabular}[c]{@{}c@{}}\#\\ of \\ arms\end{tabular}}}} &
  \multicolumn{4}{c|}{\textbf{Stiffness Modes}} \\ \cline{5-8} 
 &
  \multicolumn{1}{c|}{} &
  \multicolumn{1}{c|}{} &
  \multicolumn{1}{c|}{} &
  \multicolumn{2}{c|}{\textbf{Position}} &
  \multicolumn{2}{c|}{\textbf{Rotation}} \\ \cline{5-8} 
 &
  \multicolumn{1}{c|}{} &
  \multicolumn{1}{c|}{} &
  \multicolumn{1}{c|}{} &
  \multicolumn{1}{c|}{\textbf{low}} &
  \multicolumn{1}{c|}{\textbf{high}} &
  \multicolumn{1}{c|}{\textbf{low}} &
  \textbf{high} \\ \hline
\textbf{A} &
  \multicolumn{1}{c|}{40} &
  \multicolumn{1}{c|}{1} &
  \multicolumn{1}{c|}{1} &
  \multicolumn{1}{c|}{300} &
  \multicolumn{1}{c|}{800} &
  \multicolumn{1}{c|}{100} &
  150 \\ \hline
\textbf{B} &
  \multicolumn{1}{c|}{60} &
  \multicolumn{1}{c|}{2} &
  \multicolumn{1}{c|}{1} &
  \multicolumn{1}{c|}{800} &
  \multicolumn{1}{c|}{1200} &
  \multicolumn{1}{c|}{150} &
  300 \\ \hline
% \textbf{C} &
%   \multicolumn{1}{c|}{60} &
%   \multicolumn{1}{c|}{2} &
%   \multicolumn{1}{c|}{1} &
%   \multicolumn{1}{c|}{400} &
%   \multicolumn{1}{c|}{800} &
%   \multicolumn{1}{c|}{100} &
  % 150 \\ \hline
\textbf{C / D} &
  \multicolumn{1}{c|}{60} &
  \multicolumn{1}{c|}{3} &
  \multicolumn{1}{c|}{2} &
  \multicolumn{1}{l|}{\begin{tabular}[c]{@{}l@{}}R1: 800\\ R2: 200\end{tabular}} &
  \multicolumn{1}{l|}{\begin{tabular}[c]{@{}l@{}}1200\\800\end{tabular}} &
  \multicolumn{1}{l|}{\begin{tabular}[c]{@{}l@{}}150\\ 100\end{tabular}} &
  \multicolumn{1}{l|}{\begin{tabular}[c]{@{}l@{}}300\\ 150\end{tabular}} \\ \hline
% \textbf{D} &
%   \multicolumn{1}{c|}{60} &
%   \multicolumn{1}{c|}{3} &
%   \multicolumn{1}{c|}{2} &
%   \multicolumn{1}{l|}{\begin{tabular}[c]{@{}l@{}}R1: 800\\ R2: 200\end{tabular}} &
%   \multicolumn{1}{l|}{\begin{tabular}[c]{@{}l@{}}1200\\ 800\end{tabular}} &
%   \multicolumn{1}{l|}{\begin{tabular}[c]{@{}l@{}}150\\ 100\end{tabular}} &
%   \multicolumn{1}{l|}{\begin{tabular}[c]{@{}l@{}}300\\ 150\end{tabular}} \\ \hline
\end{tabular}
\end{table}

% \begin{itemize}[leftmargin=*]
\begin{itemize}[label={}, leftmargin=*]

\begin{figure*}[t]
    \centering
    \includegraphics[width=\textwidth]{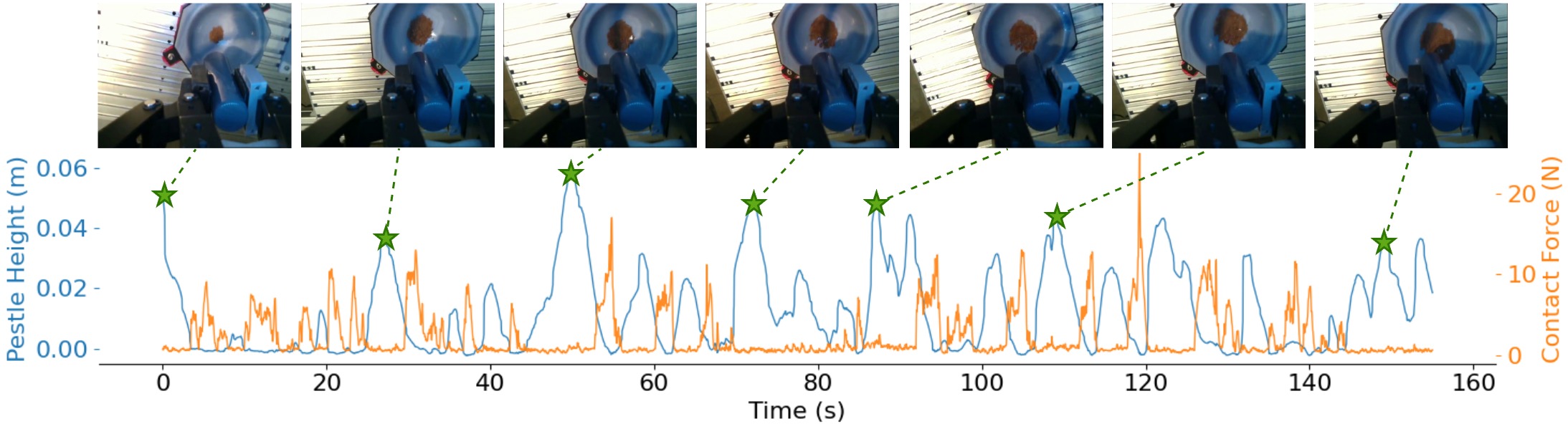}
    \caption{Powder Grinding performance by the \methodAbbr~policy. The policy imitates the demonstrated behavior of pausing every few seconds to look at the powder's state before continuing the grinding process. Position Z indicates the height of the tip of the pestle relative to the mortar.}
    \label{fig:powder_grinding_snapshots}
\end{figure*}

\begin{figure}[t]
    \centering
    \includegraphics[width=\linewidth]{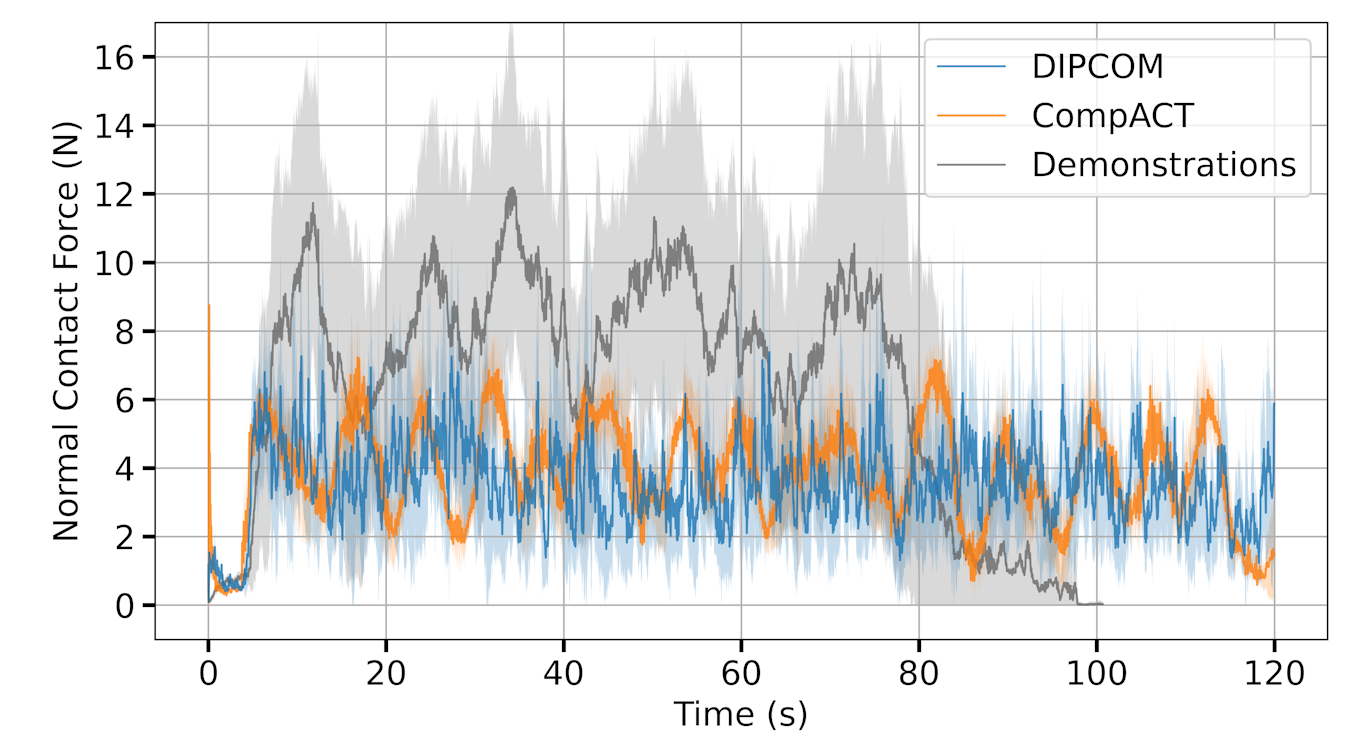}
    \caption{A - Powder Grinding: Force profile comparison between the demonstrations, the proposed method \methodAbbr, and the baseline Comp-ACT. A bold line and color-shaded area represent the average and standard deviation normal contact force, respectively.}
    \label{fig:powder_grinding_average_ft}
\end{figure}

    \item \textbf{\textit{A - Powder grinding}}: The task starts with the robot already holding the ceramic pestle and the granular powder placed at the center of the ceramic mortar. A single camera was used for this task, the one attached to the robot's wrist, as the robot would obstruct any other external view of the inside of the mortar. For this reason, the demonstrations consisted of pressing the pestle against the mortar and performing circular motions for a few seconds, then moving the robot out of the mortar to get a clear view of the state of the powder, and then repeating this process for about 80 seconds. 
    We measured the fine powder produced after each demonstration and compared it against the policy performance for this task. For practical purposes, instant coffee powder was used for experimentation as it comes in a granular form that can be ground to a finer powder. Similarly, a tea strainer was used to sift the powder.

% Please add the following required packages to your document preamble:
% \usepackage{booktabs}
\begin{table}[t]
\caption{A - Powder grinding results}
\label{table:powder_grinding_results}
\centering
\begin{tabular}{lcc}
\hline
\multicolumn{3}{c}{Percentage of fine powder produced} \\
                       & Average  & Standard Deviation \\ \hline
Human demonstrations   & 76.67\%    & 8.4\%               \\
\methodAbbr  & \textbf{55.88\%}    & 13.54\%              \\
CompACT                & 9.96\%     & 1.39\%               \\ \hline
\end{tabular}
\end{table}
    \textbf{\textit{Results}}: As shown in \Cref{table:powder_grinding_results}, our proposed method achieved the highest percentage of fine powder produced at 56 \% compared to the 10 \% achieved by Comp-ACT. Both methods applied a similar magnitude of force against the powder and mortar, as illustrated in \cref{fig:powder_grinding_average_ft}. Nevertheless, our \methodAbbr~policy obtained better results by reproducing the demonstrated behavior of moving the pestle in circular motions and periodically checking the state of the powder before repeating the grinding action, as shown in \cref{fig:powder_grinding_snapshots}. On the contrary, the Comp-ACT policy struggled to reproduce the circular motions with the pestle, mostly staying in the same place during the entire duration of each test.

\begin{figure}[t]
    \centering
    \includegraphics[width=\linewidth]{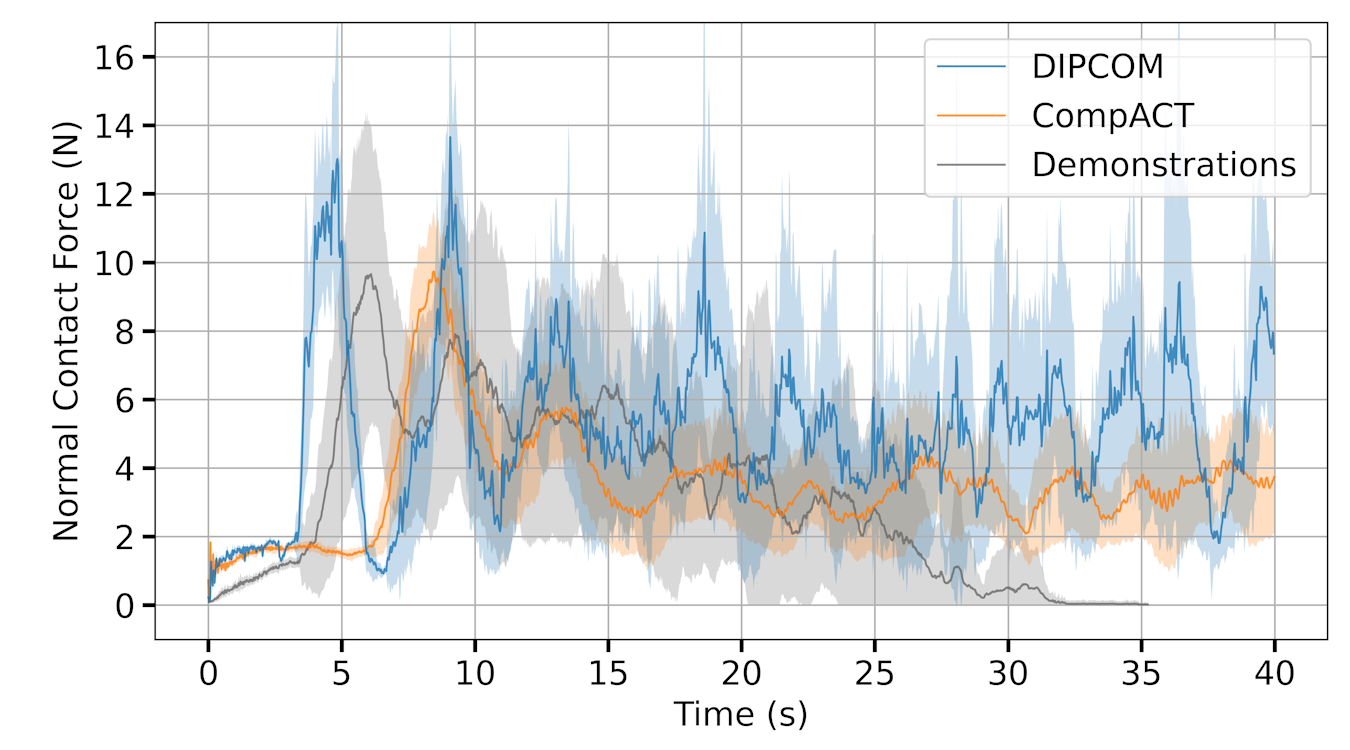}
    \caption{B - Pencil Eraser. Force profile comparison between the demonstrations, the proposed method \methodAbbr, and the baseline Comp-ACT. }
    \label{fig:pencil_eraser_ft}
\end{figure}
% Please add the following required packages to your document preamble:
% \usepackage{booktabs}
\begin{table}[t]
\centering
\caption{B - Pencil eraser task results}
\label{table:pencil_eraser_results}
\begin{tabular}{@{}lcc@{}}
\toprule
\multicolumn{1}{c}{Method} & \begin{tabular}[c]{@{}c@{}}Percentage Erased\\ Average (SD)\end{tabular} & Success Rate    \\ \midrule
Comp-ACT                    & 26.0\% (16.6\%)                                                       & 0.0\%           \\
\methodAbbr      & \textbf{77.32\% (19.48\%)}                                            & \textbf{52.3\%} \\ \bottomrule
\end{tabular}
\end{table}
    \item \textbf{\textit{B - Pencil eraser}}: 
    The task's goal is to use a rubber eraser to remove pencil markings from a notepad. The task begins with the word "OSX" written on the notepad fixed to the table while the robot already holds the rubber eraser. The robot has to gently but firmly press the eraser against the paper to remove the pencil marks without damaging the paper. During data collection, the demonstrator rubbed the eraser from right to left in a straight line, then lifted the eraser from the paper, moved back to the right side of the marks, and repeated several times until the mark was removed completely. The task was evaluated with two metrics. First, the success rate where success was defined by whether the pencil marks were completely erased or not. Second, we compute the percentage of erased pencil marks by analyzing the paper's initial and final state after each rollout. These results are reported in \Cref{table:pencil_eraser_results}.

    \textbf{\textit{Results}}: As shown in \cref{fig:pencil_eraser_ft}, the Comp-ACT policy tends to apply less overall force to the paper but struggles to completely erase the marks on the paper. On the contrary, our \methodAbbr~policy applied more force to the paper while simultaneously aiming better toward the pencil marks. As a result, our proposed method achieved a 52.3\% success rate, while Comp-ACT could not complete the task on any of the 20 rollouts. Additionally, on average, our proposed method can erase 51\% more marks than Comp-ACT.

%%%% 
    
    \textbf{\item C / D - Bimanual insertion tasks:} The task begins with one arm already holding the peg while the other arm similarly holds the mating part. The goal is to align, insert, and release the peg. The peg and hole have a tolerance of 2~mm. The peg insertion task is already challenging, but coordinating both arms to solve the task significantly increases the difficulty. Success is defined as the correct insertion and release of the peg for these tasks.

    \textbf{\textit{Results}}: As shown in \Cref{table:insertion_results}, both methods achieve a similar performance, where the round peg insertion was successful every single time, but both models struggle to solve the cuboid peg insertion 5\% of the times. Nevertheless, the behavior displayed by each method is very different. The Comp-ACT policy consistently approaches the task in the same manner. Conversely, the \methodAbbr~policy seemingly attempts different trajectories, similar to the diversity of demonstrations provided.
    Interestingly, while the demonstrators typically followed a consistent strategy of fixing the pose of the mating part with one arm before inserting the peg with the other, DIPCOM introduced motor skills not seen in the demonstrations. The DIPCOM policy continuously adjusted the supporting arm during insertion, assisting the peg-holding arm to ensure smooth and safe task execution. In contrast, Comp-ACT stuck rigidly to the demonstrated strategy. 
    
\end{itemize}

\begin{table}[t]
\centering
\caption{Success rate for bimanual insertion tasks}
\label{table:insertion_results}
\begin{tabular}{lcc}
\hline
\multicolumn{1}{c}{Task Name}   & Comp-ACT & \methodAbbr \\ \hline
C - Bimanual Round Insertion & 100\%    & 100\%                      \\ 
D - Bimanual Cuboid Insertion   & 95\%     & 95\%                       \\ \hline
\end{tabular}
\end{table}

\section{Discussion}
As mentioned above, both DIPCOM and Comp-ACT achieved similar overall outcomes in the bimanual insertion tasks, but their task execution behaviors varied notably. 
These differences became more pronounced in longer horizon tasks, such as powder grinding and pencil erasure, which demand repetitive or continuous actions based on recurring observations and don't have a linear flow. Comp-ACT initially performed well in these tasks—accurately targeting the writing during erasure and correctly approaching the powder in the grinding task. However, Comp-ACT struggled with maintaining fluid, repetitive motions, often freezing mid-task during actions like up-and-down movements for erasing or circular motions for grinding. Conversely, DIPCOM handled these repetitive tasks more flexibly, continuing with smooth, adaptive actions despite exhibiting more variance in its execution.
These results regarding the diversity of behaviors exhibited by Comp-ACT (a VAE-based method) and DIPCOM (a Diffusion-based method) agree with those reported by Jia et al.~\cite{jia2024towards} despite the difference in task state and action spaces.

Another observation was about force application during these tasks. Both policies applied force more conservatively than the human demonstrators. However, despite having a larger standard deviation, DIPCOM's diffusion-based policy better mimicked the demonstrators' force patterns. Comp-ACT, on the other hand, applied force more consistently but tended to fall into cyclic force patterns, especially in long-horizon tasks.

\subsubsection*{Limitations}
During training and fine-tuning, it was noted that DIPCOM is more sensitive to parameter settings than Comp-ACT. Diffusion-based policies, in general, require more computational resources, making them susceptible to disturbances in the robotic system's overall control frequency, which can impact their performance. This computational overhead makes DIPCOM more challenging to tune and maintain than Comp-ACT. Future work will explore the impact of hyperparameters tunning and alternative action spaces, such as the relative trajectory proposed by \cite{chi2024universal}. 

It is important to highlight that, in this study, the number of demonstrations used to train the policies was relatively small compared to other works exploring similar methods \cite{chi2023diffusion}\cite{zhaoaloha}.
% We anticipate that increasing the number of demonstrations will enhance performance, particularly for DIPCOM, by further refining its skill set. 
An interesting future research avenue is to explore the effects of large-scale datasets for compliant manipulation tasks aiming to improve policy generalization across extended task variations. For instance, instead of training a policy for each specific powder grinding task, we aim to develop a single policy that can perform across different grinding tasks, broadening its applicability and robustness.

\section{Conclusions}

This study introduced \methodName~(\methodAbbr), a diffusion-based framework designed for compliant control tasks, particularly for rigid robots. Our approach demonstrates how diffusion policies effectively capture the multimodality of the data, predicting Cartesian end-effector poses while adjusting the arm's stiffness to apply the required contact forces. We provide guidelines for implementing diffusion policies for compliant control and outline best practices for optimizing performance. Through extensive experimental evaluations, we showcased \methodAbbr's strengths in various contact-rich tasks, highlighting its advantages over previous methods. Moving forward, we aim to explore the policy's ability to generalize across different task variations by expanding the dataset. We plan to enhance force processing and refine the policy architecture to improve force-aware inference.

% \section*{Acknowledgment}

\bibliographystyle{IEEEtran}
\bibliography{ref}

\end{document}